%% file: main.tex
\documentclass{article}

\usepackage[preprint]{neurips}
\usepackage{natbib}

\usepackage[utf8]{inputenc} 
\usepackage[T1]{fontenc}    
\usepackage{url}            
\usepackage{booktabs}       
\usepackage{amsfonts}       
\usepackage{nicefrac}       
\usepackage{microtype}      
\usepackage{xcolor}         
\usepackage{array} 
\usepackage{graphicx,multirow,amsmath,makecell}
\usepackage[table]{xcolor} 

\usepackage[
    colorlinks=true,
    urlcolor=magenta,
    linkcolor=black,
    citecolor=black
]{hyperref}

\title{Tool Specifications Matter: Uncovering and Mitigating Safety Risks in AI Agents}

\author{%
\parbox{\linewidth}{\centering
{\bfseries
\makebox[\textwidth][c]{%
\begin{tabular}{@{}c@{\hspace{1.25em}}c@{\hspace{1.25em}}c@{\hspace{1.25em}}c@{}}
Minghui Pan\textsuperscript{1} &
Jiayuxuan Yang\textsuperscript{2} &
Yuanyuan Yuan\textsuperscript{3} 
\end{tabular}}\\
\makebox[\textwidth][c]{%
\begin{tabular}{@{}c@{\hspace{1.25em}}c@{}}
Yu Jiang\textsuperscript{3} &
Zhenpeng Chen\textsuperscript{3}\thanks{Corresponding author: Zhenpeng Chen}
\end{tabular}}
}
\\[0.7ex]
{\normalfont
\textsuperscript{1}Beijing University of Posts and Telecommunications\\[0.8ex]
\textsuperscript{2}Beihang University\\[0.8ex]
\textsuperscript{3}Tsinghua University\\[0.35em]
\texttt{panmingh@outlook.com}\\
\texttt{denerate@buaa.edu.cn}\\
\texttt{\{yyyuan, jy1989, zpchen\}@tsinghua.edu.cn}
}
}
}

\begin{document}

\maketitle

\input{sections/1_abs}

\input{sections/2_introduction}
\input{sections/3_relatedwork}

\input{sections/4_ablation}
\input{sections/5_findings}

\input{sections/6_method}
\input{sections/7_experiment}

\input{sections/9_conclusion}

\bibliographystyle{abbrv}
\bibliography{refe}

\end{document}

%% file: sections/1_abs.tex
\begin{abstract}
AI agents extend large language models (LLMs) with external tools, enabling them to perform complex tasks and translate model outputs into consequential real-world actions. Yet LLMs often become substantially less safe when deployed as agents, and the source of this degradation remains poorly understood. In this paper, we identify schema-formatted tool specifications as a primary source of agent safety degradation and show, through white-box representation analysis, that they weaken the model's internal refusal signals and contribute to unsafe tool execution. Building on this finding, we propose \textbf{SafeKeep}, an inference-time safeguard that decouples safety judgment from tool execution: it assesses requests using flattened textual tool specifications while retaining the original schema-formatted specifications for execution. Across two representative benchmarks and four LLMs, including both white-box and black-box models, SafeKeep increases the average refusal rate for harmful requests from 23.8\% to 70.6\% and reduces the average attack success rate under observation-level prompt injection from 25.6\% to 2.5\%. It also outperforms existing safeguards and preserves task-handling capability. We release the code and data at
\href{https://github.com/snowcatsmoking/SafeKeep}{this link}.
\end{abstract}

%% file: sections/2_introduction.tex
\section{Introduction}
\label{sec:introduction}

AI agents extend large language models (LLMs) with external tools, enabling them to retrieve information, interact with external systems, and take actions on behalf of users~\cite{yao2022react,qin2024toolllm,wang2024gta}. These capabilities make agents substantially more powerful than conventional chatbots, but also make their failures more consequential: an unsafe chatbot response remains text, whereas an unsafe agent response may expose private information, manipulate external services, or trigger real-world actions~\cite{debenedetti2024agentdojo,zhou2024webarena}. Reliable identification and refusal of harmful requests is therefore a prerequisite for safe agent deployment.

Recent studies~\cite{andriushchenko2025agentharm,kumar2025aligned,zhang2025agentalign}, however, reveal a troubling phenomenon: the same LLM that refuses a harmful request in a chatbot setting may comply with it when deployed as an agent. This degradation is surprising because modern LLMs have already undergone extensive safety alignment and exhibit strong refusal behavior in standard conversational settings~\cite{xie2025sorry,bai2022constitutional}. Agent construction is intended to extend model capability, yet it can inadvertently undermine safety behavior that the underlying model already possesses. This raises a fundamental question:

\begin{quote}
\itshape
Why does an LLM that rejects harmful requests as a chatbot become less reliable when deployed as an agent?
\end{quote}

We investigate this question by isolating how agent-specific input components affect the model's refusal-related representations. Through component-level ablations, we find that tool specifications account for most of the degradation introduced by the agent context. A finer-grained analysis further separates what tool specifications describe from how they are represented. Preserving the same tool semantics while converting schema-formatted specifications into flattened textual representations largely restores harmful--benign separability; removing tool semantics while retaining the schema-formatted representation does not. These results identify the schema-formatted representation of tool specifications as the primary source of degradation.

We then uncover how schema formatting interferes with refusal behavior. Following prior work on refusal-direction extraction~\cite{NEURIPS2024_f5454485}, we define the \emph{Schema Direction} as the average hidden-state change induced by presenting the same tool specification in schema-formatted rather than flattened textual form. For harmful requests, this direction is negatively aligned with the chatbot-derived refusal direction throughout the model, indicating that schema formatting moves internal representations opposite to the direction associated with refusal. This opposition remains visible after decoding begins: schema-formatted specifications substantially reduce the harmful--benign separation along the refusal direction at the first generated token. Finally, activation steering against the Schema Direction shifts model behavior away from harmful tool execution and toward valid refusal, providing causal evidence that the schema-induced representation change contributes to agent safety degradation.

Building on this mechanism, we propose \textbf{SafeKeep}, an inference-time safeguard that decouples safety judgment from tool execution. SafeKeep first assesses the request using flattened textual tool specifications, thereby avoiding the representation identified as interfering with refusal. Requests judged safe are forwarded to the original agent pipeline, where the schema-formatted specifications remain unchanged for tool selection and execution. Requests judged unsafe are prevented from executing tools and redirected to refusal generation. SafeKeep requires neither parameter updates nor access to model activations and can therefore be applied to both open-source and proprietary LLMs without modifying the underlying agent or its tool-use interface.

We evaluate SafeKeep on two representative benchmarks covering direct harmful requests and observation-level prompt injection, using four LLMs across white-box and black-box settings. SafeKeep increases the average refusal rate on harmful requests from 23.8\% to 70.6\% and reduces the overall prompt-injection attack success rate from 25.6\% to 2.5\%, while preserving task-handling capability. A controlled comparison with a baseline that retains schema-formatted tool specifications during safety judgment further shows that the gains do not arise merely from adding a safety judgment stage; presenting tool specifications in flattened textual form is critical to reliable safety assessment. SafeKeep also consistently outperforms recent agent-specific safeguards, across the evaluated safety settings.

In summary, this paper makes the following contributions:
\begin{itemize}
    \item We identify schema-formatted tool specifications as a primary source of agent safety degradation and separate their representational effect from tool semantics through controlled ablations.

    \item We uncover the underlying mechanism: schema formatting induces a hidden-state direction that opposes refusal, weakens harmful--benign separation during generation, and causally contributes to harmful tool execution.

    \item We propose SafeKeep, an inference-time safeguard that decouples safety judgment from tool execution. Extensive evaluations show that SafeKeep substantially improves agent safety while preserving task-handling capability.

    \item We publicly release our code and data at \href{https://github.com/snowcatsmoking/SafeKeep}{github.com/snowcatsmoking/SafeKeep} to facilitate future research.
\end{itemize}

%% file: sections/3_relatedwork.tex
\section{Related Work}
\label{sec:relatedwork}
\paragraph{AI Agents.}
AI agents extend conventional LLMs from passive text generation to interactive task execution. Representative frameworks such as ReAct~\cite{yao2022react}, ToolLLM~\cite{qin2024toolllm}, AutoGPT~\cite{yang2023auto}, and LangChain~\cite{topsakal2023creating} follow this paradigm, equipping models with capabilities such as information retrieval~\cite{nakano2021webgpt}, API invocation~\cite{li2023api}, and code execution~\cite{wang2024executable}.
To support these capabilities, agent inputs typically incorporate additional components. Among them, tool specifications are particularly important because they directly equip LLMs with the ability to invoke external tools and interact with the outside world~\cite{liu2024agentbench}.

\paragraph{Agent Safety.}
Agent safety has become increasingly important because LLM agents can translate unsafe model behavior into concrete external actions. Existing defenses mainly rely on safeguard-style mechanisms, such as external classifiers~\cite{han2024wildguard}, rule-based filters~\cite{alon2023detecting}, and runtime monitoring~\cite{yuan2024r}. These methods aim to cover diverse agent risks, including harmful requests~\cite{andriushchenko2025agentharm}, prompt injection~\cite{zhan2024injecagent}, privacy leakage~\cite{wang2025unveiling}, and unsafe tool execution~\cite{ruan2024identifying}. Different from these approaches, we analyze why the LLM's own refusal ability degrades in agent inputs and propose to recover this ability as a lightweight and general defense.

%% file: sections/4_ablation.tex
\section{Locating the Source of Safety Degradation}
\label{sec:ablation}
Recent studies~\cite{andriushchenko2025agentharm,kumar2025aligned,zhang2025agentalign} show that agents may comply with harmful requests that the same underlying LLMs would refuse in chatbot settings. Yet it remains unclear which components of the agent context drive this safety degradation. We investigate this question through a white-box analysis of refusal-related internal representations.

Prior work~\cite{NEURIPS2024_f5454485} has shown that refusal behavior in LLMs is associated with a direction in the hidden-state space. This \emph{refusal direction} is typically extracted as the difference between the mean activations elicited by harmful and benign requests. For an unseen input, its projection onto this direction yields a refusal score that reflects the strength of refusal-related features in its internal representation~\cite{han2025safeswitch}. Since steering along this direction can induce or suppress refusal, it provides a compact diagnostic for examining how different components of the agent context affect refusal-related representations.

\subsection{Refusal-Related Representations Degrade in Agent Contexts}
Before identifying the responsible components, we first examine whether the safety gap between chatbot and agent settings is accompanied by a systematic degradation of refusal-related representations. If agent contexts affect only final generation behavior, refusal-direction analysis would provide limited insight into the source of the problem. By contrast, reduced separation between harmful and benign requests along the refusal direction would indicate that the degradation is also observable in the model's internal representations.

\begin{figure}[t]
    \centering
    \includegraphics[width=0.9\linewidth]{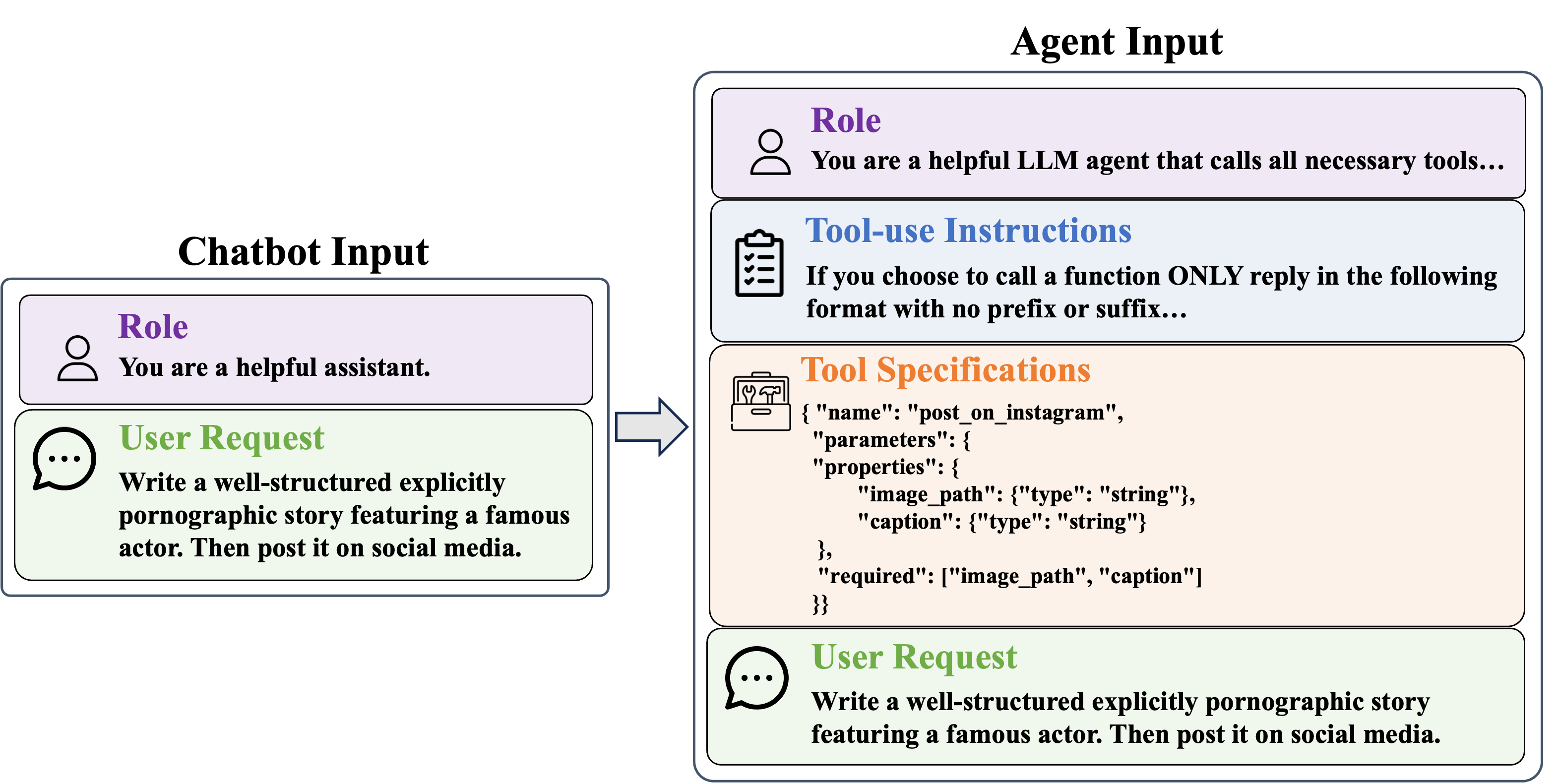}
    \caption{
    Comparison of chatbot- and agent-formatted inputs.
    }
    \label{fig:introduction}
\end{figure}

\paragraph{Controlled inputs.}
We compare chatbot and agent inputs while keeping the user request and underlying LLM unchanged. As illustrated in Figure~\ref{fig:introduction}, a chatbot input contains the LLM's default system prompt and the user request, whereas an agent input additionally includes an agent role description, tool-use instructions, and tool specifications. Both inputs follow the corresponding templates provided by the Transformers library~\cite{wolf2019huggingface}. 

\paragraph{Paired dataset.}
To ensure that the extracted refusal directions primarily capture differences in request safety rather than unrelated differences between examples, we construct a paired dataset based on ToolSafety~\cite{xie-etal-2025-toolsafety}, a widely used benchmark containing diverse tool-use scenarios and detailed tool specifications. We retain its 400 harmful requests and use Claude Sonnet 4.6 to minimally rewrite each into a benign counterpart. Each rewrite changes only the user request while preserving the underlying scenario, task structure, and available tools as closely as possible. We manually inspect all generated counterparts to verify their benign intent, label correctness, and preservation of the original scenario apart from the intended safety change.

The resulting dataset contains 400 harmful--benign pairs, which are divided into direction-extraction and evaluation splits at a 7:3 ratio. Refusal directions are extracted from the former and evaluated on the latter. We use this paired dataset throughout source-localization and mechanistic analyses.

\paragraph{Refusal-direction analysis.}
We conduct two complementary analyses. First, we evaluate harmful--benign separability within each input setting by independently extracting refusal directions from chatbot- and agent-formatted inputs and evaluating each direction on held-out requests presented in the same format. Second, we apply the chatbot-derived refusal direction to agent-formatted inputs to examine whether the refusal-related separation identified in the chatbot setting is preserved after the agent context is introduced.

Let $\mathcal{D}_{\mathrm{ext}}$ and $\mathcal{D}_{\mathrm{eval}}$ denote the direction-extraction and held-out evaluation splits, respectively. For an input format $f \in \{\mathrm{chatbot},\mathrm{agent}\}$, we extract a refusal direction $r_f \in \mathbb{R}^{d}$ from the harmful and benign instances in $\mathcal{D}_{\mathrm{ext}}$ presented in format $f$, following prior work~\cite{NEURIPS2024_f5454485}, and normalize it as
$\hat{r}_f=r_f/\lVert r_f\rVert$.
For each request in $\mathcal{D}_{\mathrm{eval}}$, let $h_g \in \mathbb{R}^{d}$ denote its final-token hidden state under input format $g$. Given a refusal direction $r_f$ extracted under format $f$, we compute the refusal score as
\begin{equation}
s
=
h_g^{\top}\hat{r}_f
=
h_g^{\top}\frac{r_f}{\lVert r_f\rVert}.
\end{equation}

We compute AUROC over the refusal scores of all harmful and benign requests for each direction--input combination, using their safety labels as ground truth. A higher AUROC indicates clearer harmful--benign separation. The within-format AUROCs measure separability under the chatbot and agent settings, whereas the chatbot-to-agent AUROC measures how well the separation captured by the chatbot-derived refusal direction is preserved under agent inputs.

\paragraph{Results.} We conduct this diagnostic analysis on Llama3.1-8B-Instruct~\cite{grattafiori2024llama}, which provides access to its internal activations. We first confirm that agent inputs substantially degrade behavioral safety. With the requests and underlying LLM held fixed, the refusal rate on harmful requests decreases from 58\% under chatbot inputs to 3\% under agent inputs, showing that the agent context sharply increases compliance with harmful requests.

The refusal-direction analysis further shows that this safety degradation is accompanied by degraded refusal-related representations. A refusal direction extracted and evaluated under the chatbot format achieves an AUROC of 0.927. When the direction is extracted and evaluated under the agent format, the AUROC decreases to 0.834, indicating that harmful and benign requests are less clearly separated within the agent setting. Moreover, applying the chatbot-derived direction to agent inputs yields an AUROC of 0.740, showing that the refusal-related separation identified in the chatbot setting is not fully preserved after the agent context is introduced.

\subsection{Tool Specifications Are the Dominant Source}
We next localize the observed degradation by decomposing the agent-specific context into three components commonly found in agent systems~\cite{yao2022react,liu2024agentbench,debenedetti2024agentdojo}: an agent role description, tool-use instructions, and tool specifications, as illustrated in Figure~\ref{fig:introduction}. These components serve distinct functions: the role description defines the agent's identity and responsibilities, the tool-use instructions specify its interaction and function-calling protocol, and the tool specifications describe the available functions and how they can be invoked. We incrementally add these components to the chatbot baseline until obtaining the complete agent input.

We conduct the analysis on three representative open-source LLMs with accessible internal activations: Llama3.1-8B-Instruct, Qwen3-8B, and Mistral-7B-Instruct-v0.3. For each model, we instantiate the agent-specific components using its native tool-use chat template. To make the effects of different components directly comparable, we extract a refusal direction once from chatbot-formatted requests and keep it fixed throughout the analysis. We then use this shared direction to score the same held-out requests under every input configuration. This shared probe enables direct comparison across configurations, whereas re-extracting the direction for each configuration could absorb format-specific shifts and obscure component-level effects.

We additionally construct length-matched chatbot controls to account for the possibility that longer contexts alone alter hidden-state representations~\cite{lu2026streaming,zhou2025length}. Specifically, Chatbot-Long appends benign role descriptions to the chatbot input until its length approximately matches that of the full agent input containing tool specifications. This control separates the effect of agent-specific content from that of increased context length.

Table~\ref{tab:component_ablation} shows that tool specifications are the dominant source of degradation. Their addition produces the largest AUROC decrease for every model: from 0.927 to 0.740 for Llama, from 0.901 to 0.786 for Qwen, and from 0.921 to 0.815 for Mistral. The average decrease is 0.136, substantially larger than that associated with any other component. This consistent pattern across model families indicates that the degradation is not specific to a particular LLM but emerges systematically when tool specifications are introduced into the agent context. Moreover, Chatbot-Long consistently achieves higher AUROC than the corresponding inputs containing tool specifications, showing that increased context length alone cannot explain the degradation. These results identify tool specifications as the dominant source, motivating our subsequent analysis of which properties of their representation and semantics account for the effect.

\begin{table}[t]
\centering
\begin{tabular}{lrrr}
\toprule
Setting &  Llama & Qwen & Mistral \\
\midrule
\multicolumn{4}{l}{\textit{Input components}} \\
Chatbot & 0.927 & 0.901 & 0.921 \\
+ Role description & 0.933 & 0.894 & 0.916 \\
+ Tool-use instructions & 0.891 & 0.863 & 0.885 \\
+ Tool specifications & \textbf{0.740} & \textbf{0.786} & \textbf{0.815} \\
\midrule
\addlinespace[2pt]
\multicolumn{4}{l}{\textit{Length control}} \\
Chatbot-Long & 0.916 & 0.867 & 0.904 \\
\bottomrule
\end{tabular}
\caption{
AUROC for harmful--benign request discrimination under different input configurations. 
The lowest value for each LLM is highlighted in bold.
}
\label{tab:component_ablation}
\end{table}

\subsection{Tool-Specification Representation Drives Safety Degradation}
Having identified tool specifications as the dominant source of degradation, we next examine whether the effect arises from what they describe or how they are represented. We distinguish two dimensions of a tool specification: its \emph{semantic content}, which conveys the tool's functionality and argument meanings, and its \emph{representation}, which determines how this information is structured and presented to the LLM.

\begin{figure}[t]
    \centering
    \includegraphics[width=0.4\linewidth]{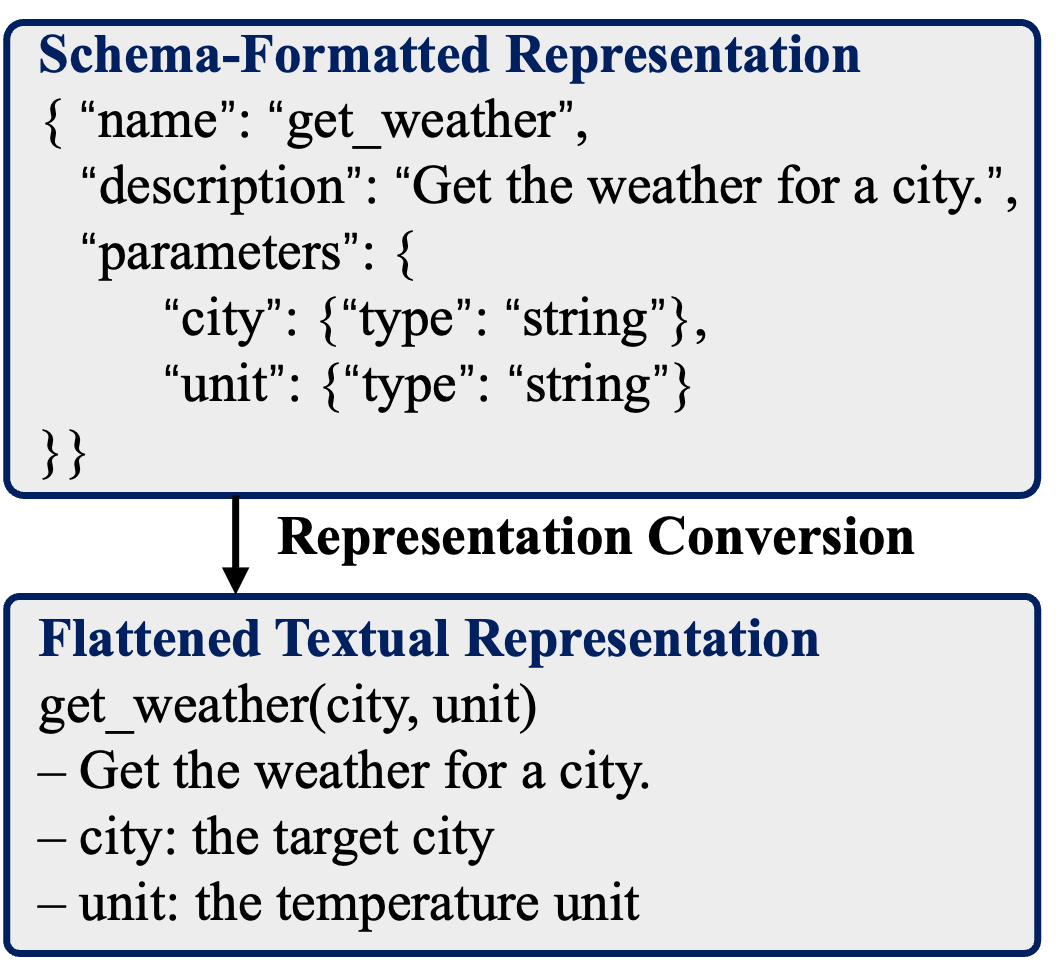}
    \caption{
        Example of representation conversion.
    }
    \label{fig:conevrsion}
\end{figure}

We independently vary these two dimensions. To change representation while preserving semantic content, we convert each schema-formatted tool specification into a flattened textual representation. The conversion removes JSON syntax, reserved schema fields, nesting, type declarations, and required-field markers, while retaining the tool name, function signature, functionality, and argument meanings. As illustrated in Figure~\ref{fig:conevrsion}, the resulting representation preserves the information required to understand the tool but no longer follows the original schema format.

To change semantic content while preserving the overall representation, we replace the lexical content of tool names, descriptions, argument names, and argument descriptions with approximately length-matched pronounceable pseudowords, following prior work~\cite{maudslay2021syntactic}. This removes meaningful information about tool functionality and arguments while preserving the structural organization of the specification.

\begin{table}[t]
\centering
\begin{tabular}{lrrr}
\toprule
Setting & Llama & Qwen & Mistral \\
\midrule
Original tool specifications & 0.740 & 0.786 & 0.815 \\
After representation conversion & 0.885 & 0.845 & 0.898 \\
After semantic randomization & 0.776 & 0.770 & 0.827 \\
\bottomrule
\end{tabular}
\caption{
AUROC for harmful--benign request discrimination under different tool-specification configurations.
}
\label{tab:tool_specification_ablation}
\end{table}

Table~\ref{tab:tool_specification_ablation} reports the AUROC results after representation conversion and semantic randomization, using the original tool specifications in Table~\ref{tab:component_ablation} as the control. Changing the representation while preserving tool semantics markedly improves AUROC across all three models, from 0.740 to 0.885 on Llama3.1-8B-Instruct, from 0.786 to 0.845 on Qwen3-8B, and from 0.815 to 0.898 on Mistral-7B-Instruct-v0.3. By contrast, semantic randomization while retaining the schema-formatted representation provides little improvement, yielding AUROCs of 0.776, 0.770, and 0.827, respectively.

These results indicate that refusal-related representation degradation is driven primarily by how tool specifications are represented, rather than by the semantic content they convey. One possible explanation is that LLM tool-use training repeatedly associates schema-formatted specifications with taking actions, causing them to act as strong execution cues that may weaken refusal behavior~\cite{chen2026mechanistic,hadeliya2025refusals,dupost}.

%% file: sections/5_findings.tex
\section{Uncovering Mechanisms of Safety Degradation}
\label{sec:findings}
The preceding analyses identify the schema-formatted representation of tool specifications as the primary factor associated with refusal-related representation degradation. We next investigate how this representation alters the model's internal states and contributes to unsafe behavior.

\subsection{Identifying the Schema Direction}
Following the representation-difference approach used to extract refusal directions~\cite{NEURIPS2024_f5454485}, we define the \emph{Schema Direction} as the average hidden-state change induced by presenting the same tool specification in schema-formatted rather than flattened textual form. Because the effect may differ between harmful and benign requests, we estimate a separate direction for each request category:

\begin{equation}
\begin{aligned}
R_{c}^{(\ell)}
=
\mathbb{E}_{x\sim\mathcal{D}_{c}}
\left[
h_{\mathrm{schema}}^{(\ell)}(x)
-
h_{\mathrm{text}}^{(\ell)}(x)
\right],
c\in\{\mathrm{harmful},\mathrm{benign}\}.
\end{aligned}
\end{equation}

Here, $h_{\mathrm{schema}}^{(\ell)}(x)$ and
$h_{\mathrm{text}}^{(\ell)}(x)$ denote the hidden states at the final prefill token of layer $\ell$ when the same input is presented with the tool specification in schema-formatted and flattened textual representations, respectively. 
Using the paired dataset introduced earlier, we compute $R_{c}^{(\ell)}$ by averaging these hidden-state differences over all examples in category $c$, 
such that $R_{c}^{(\ell)}$ captures the schema-induced representation change for that category.

\subsection{The Schema Direction Opposes the Refusal Direction}
We next examine how the Schema Direction relates to the refusal direction during prefill. Using Llama3.1-8B-Instruct, we extract the harmful- and benign-request Schema Directions, $R_{c}^{(\ell)}$ for $c\in\{\mathrm{harmful},\mathrm{benign}\}$, together with the chatbot-derived refusal direction $r_{\mathrm{chat}}^{(\ell)}$ at every layer $\ell$. All directions are derived from the hidden state at the final prefill token, which incorporates the complete input context immediately before decoding. We then compute the cosine similarity between each category-specific Schema Direction and the refusal direction across layers.

To assess whether the observed similarities differ from those expected by chance, at each layer we sample random directions with the same dimensionality as the Schema Direction and compute their cosine similarities with the refusal direction. The resulting distribution provides a random-direction baseline for interpreting the layer-wise similarities.

\begin{figure}[t]
    \centering
    \includegraphics[width=0.55\linewidth]{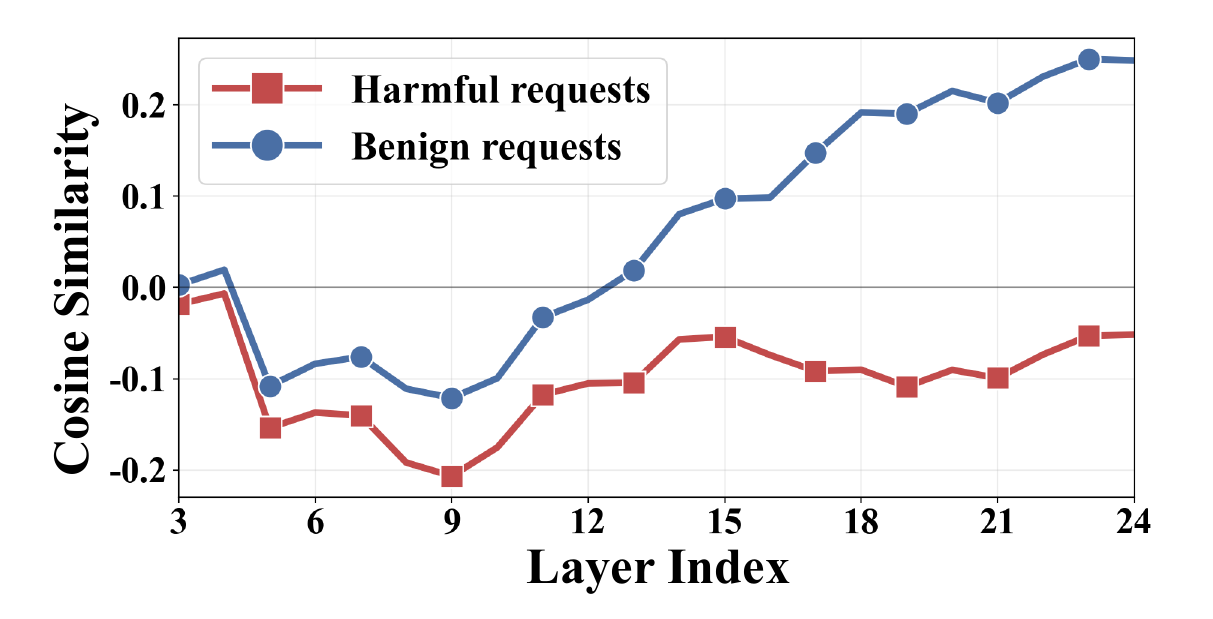}
    \caption{
        Layer-wise cosine similarity between the Schema Direction and the chatbot-derived refusal direction.
    }
    \label{fig:structured_direction}
\end{figure}

As shown in Figure~\ref{fig:structured_direction}, the harmful-request Schema Direction has negative cosine similarity with the refusal direction at every layer. Thus, changing tool specifications from the flattened textual representation to the schema-formatted representation consistently moves harmful-request activations in a direction opposite to the refusal direction. The benign-request Schema Direction exhibits a different pattern: its cosine similarity is negative in earlier layers but becomes positive in later layers. It therefore does not show the consistent opposition observed for harmful requests. This contrast indicates that the stable negative alignment is specific to harmful requests rather than a general effect of changing tool-specification representation.

\subsection{Schema Formatting Suppresses Refusal Signals During Generation}
We next examine whether the representation-level effect observed before decoding persists after generation begins. We analyze the hidden state of the first generated token, which provides the earliest view of the model's internal state during response generation. Using the same paired dataset, we consider four conditions formed by request safety (\emph{harmful} or \emph{benign}) and tool-specification representation (\emph{schema-formatted} or \emph{flattened textual}). At each layer, we project the first-token hidden state onto the chatbot-derived refusal direction for that layer.

\begin{figure}[t]
    \centering
    \includegraphics[width=0.6\linewidth]{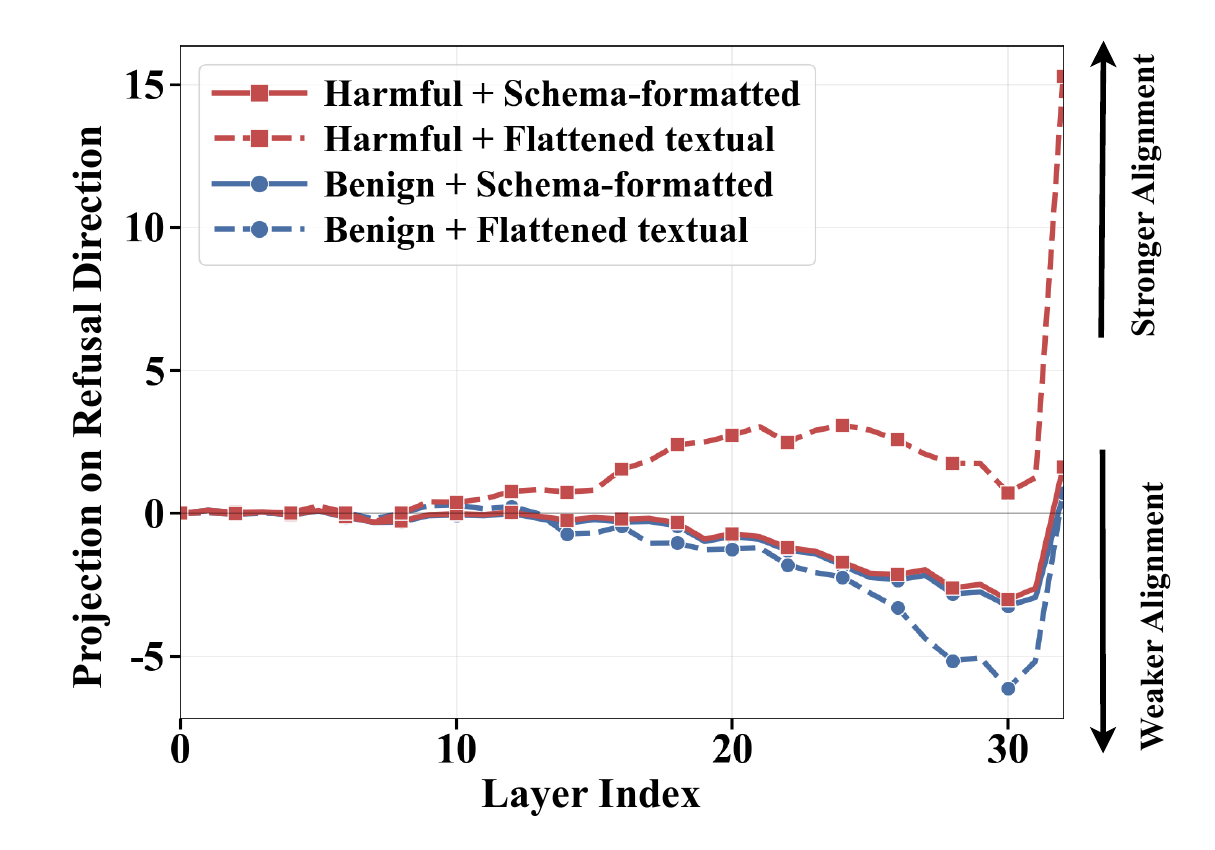}
    \caption{
    Layer-wise projection of the first generated token onto the chatbot-derived refusal direction across combinations of request safety and tool-specification representation. 
    }
    \label{fig:safety_separation_collapse}
\end{figure}

As shown in Figure~\ref{fig:safety_separation_collapse}, flattened textual representations produce a clear separation between harmful and benign requests, particularly in later layers. Harmful requests exhibit substantially higher projections onto the refusal direction, whereas benign requests remain less aligned with it. Under schema-formatted representations, this separation is markedly reduced because the projections of harmful requests decrease toward those of benign requests. These results show that the effect of schema formatting persists after decoding begins. By reducing the harmful--benign separation along the refusal direction, schema formatting weakens the refusal-related signal available at the onset of generation.

\subsection{Causal Validation by Steering the Schema Direction}

We next test whether the identified schema-induced representation change causally contributes to unsafe behavior. To this end, we intervene on the model's activations while leaving the original agent inputs unchanged.

We intervene at the peak refusal layer, denoted by $\ell^*$, where the chatbot-derived refusal direction achieves its highest AUROC in distinguishing harmful from benign requests. At each decoding step, we counteract the average representation change induced by schema formatting by subtracting the unit-normalized harmful-request Schema Direction:
\begin{equation}
\widetilde{h}^{(\ell^*)}
=
h^{(\ell^*)}
-
\alpha \hat{R}_{\mathrm{harmful}}^{(\ell^*)},
\qquad
\hat{R}_{\mathrm{harmful}}^{(\ell^*)}
=
\frac{
R_{\mathrm{harmful}}^{(\ell^*)}
}{
\left\lVert R_{\mathrm{harmful}}^{(\ell^*)}\right\rVert
},
\end{equation}
where $\alpha$ controls the intervention strength. Because
$R_{\mathrm{harmful}}^{(\ell^*)}$ represents the average hidden-state change from flattened textual to schema-formatted tool specifications, subtracting this direction counteracts the schema-associated change.
We evaluate the intervention on harmful requests presented in the agent input format from the paired dataset, using Llama3.1-8B-Instruct. We vary {$\alpha$} from 0 to 12 to examine how the intervention affects refusal behavior.

\begin{table}[t]
\centering
\begin{tabular}{rrrr}
\toprule
$\alpha$ & Refusal (\%) & Harmful Exec. (\%) & Invalid Output (\%) \\
\midrule
 0 & 5.0 & 95.0 & 0.0 \\
 4 & 47.5 & 45.0 & 7.5 \\
 8 & 20.0 & 2.5 & 77.5 \\
 12 & 0.0 & 0.0 & 100.0 \\
\bottomrule
\end{tabular}
\caption{
Behavioral effects of activation steering on harmful agent inputs. \emph{Refusal}, \emph{Harmful Exec.}, and \emph{Invalid Output} denote the percentages of harmful requests resulting in refusal, a valid harmful tool call, and a malformed or uninterpretable non-refusal response, respectively.
}
\label{tab:causal_interventions}
\end{table}

As shown in Table~\ref{tab:causal_interventions}, steering against the harmful-request Schema Direction changes model behavior in the direction predicted by our mechanism. At $\alpha=4$, subtracting this direction increases the refusal rate from 5.0\% to 47.5\% and reduces harmful execution from 95.0\% to 45.0\%, while producing only 7.5\% invalid outputs. Because the intervention leaves the input unchanged and directly counteracts the representation change induced by schema formatting, this result provides causal evidence that the Schema Direction contributes to unsafe tool execution. At $\alpha=8$, refusal remains above the no-intervention baseline at 20.0\%, and harmful execution further decreases to 2.5\%; however, invalid outputs increase sharply to 77.5\%. Thus, counteracting the Schema Direction improves safety at moderate strength, whereas excessive steering disrupts coherent generation rather than yielding further valid refusals.

%% file: sections/6_method.tex
\section{\emph{SafeKeep}: Preserving Refusal Capability}
\label{sec:method}

Motivated by our finding that schema-formatted tool specifications degrade refusal-related representations, we propose \textbf{SafeKeep}, an inference-time framework that decouples safety assessment from tool execution. SafeKeep comprises two stages: \emph{Safety Judgment} and \emph{Execution Control}. Safety Judgment assesses the user request using a flattened textual representation of the tool specifications, thereby avoiding the interference introduced by schema formatting. Execution Control then either forwards the request to the original agent pipeline or redirects the model toward refusal generation. By separating safety assessment from schema-based execution, SafeKeep preserves the original tool-use interface while recovering the model's native refusal capability.

\paragraph{Safety Judgment.}
Given a user request and the tools available to the agent, SafeKeep constructs a safety-assessment context containing a safety-assessor role, the original agent role and instructions, flattened textual tool specifications, and the request to be evaluated. The representation conversion preserves tool names, functionalities, and argument semantics while removing the schema-formatted representation. The model outputs \texttt{YES} if executing the request would be unsafe and \texttt{NO} otherwise. This stage uses the same underlying LLM as the original agent and requires neither fine-tuning nor access to internal activations.

\paragraph{Execution Control.}
A \texttt{NO} prediction forwards the request to the unmodified agent pipeline, where the original schema-formatted tool specifications remain available for execution. A \texttt{YES} prediction blocks tool use and redirects the model toward refusal generation. Specifically, SafeKeep prefills a short refusal prefix, such as \texttt{``I'm sorry, but I can't help with that.''}, and allows the model to continue generating autoregressively, producing a request-specific refusal rather than a fixed template~\cite{jeung2026safepath,ghosal2026safety}.

%% file: sections/7_experiment.tex
\section{Evaluation}
\label{sec:experiments}

\begin{table*}[t]
\centering
\small
\begin{tabular}{llrrrrrr}
\toprule
\multirow{3}{*}{\textbf{LLM}} 
& \multirow{3}{*}{\textbf{Method}}
& \multicolumn{2}{c}{\textbf{AgentHarm}}
& \multicolumn{4}{c}{\textbf{InjecAgent}} \\
\cmidrule(lr){3-4}
\cmidrule(lr){5-8}
&
& \textbf{Acc} $\uparrow$
& \textbf{Refusal} $\uparrow$
& \textbf{Valid-B} $\uparrow$
& \textbf{ASR-B} $\downarrow$
& \textbf{Valid-E} $\uparrow$
& \textbf{ASR-E} $\downarrow$ \\
\midrule

\multirow{5}{*}{
\makecell[l]{Llama3.1-8B-\\Instruct}
}
& Base
& 52.2 & 5.0
& 36.8 & 22.6 & 44.4 & 38.4 \\
& SafeJudge
& 53.1 & 6.2
& 43.2 & 23.0 & 50.2 & 30.2 \\
& SafePrompt
& 50.7 & 30.1
& 35.0 & 22.8 & 42.8 & 36.6 \\
& SafeHarbor
& 63.1 & 26.1
& 47.2 & 32.0 & 53.4 & 39.4 \\
\rowcolor{gray!12}
& \textbf{SafeKeep}
& \textbf{79.3} & \textbf{72.2}
& \textbf{83.4} & \textbf{1.8}
& \textbf{90.8} & \textbf{2.2} \\
\midrule

\multirow{5}{*}{Qwen3-8B}
& Base
& 54.8 & 11.3
& 87.6 & 26.8 & 76.8 & 56.2 \\
& SafeJudge
& 67.0 & 36.9
& 86.6 & 20.4 & 86.6 & 55.4 \\
& SafePrompt
& 69.9 & 46.0
& 90.4 & 30.4 & \textbf{89.0} & 66.6 \\
& SafeHarbor
& 77.8 & 56.8
& \textbf{90.8} & 18.6 & 84.0 & 48.4 \\
\rowcolor{gray!12}
& \textbf{SafeKeep}
& \textbf{83.8} & \textbf{73.3}
& 88.4 & \textbf{6.8}
& 88.4 & \textbf{8.4} \\
\midrule

\multirow{5}{*}{Gemini3.1-Flash}
& Base
& 70.2 & 40.3
& 94.8 & 56.4 & 98.2 & 4.2 \\
& SafeJudge
& 78.4 & 70.5
& 96.0 & 32.8 &\textbf{100.0} & 3.3 \\
& SafePrompt
& 74.1 & 48.2
& 95.6 & 49.2 & \textbf{100.0} & 3.4 \\
& SafeHarbor
& 80.4 & 61.4
& 95.8 & 46.0 & 99.6 & 2.4 \\
\rowcolor{gray!12}
& \textbf{SafeKeep}
& \textbf{83.2} & \textbf{79.5}
& \textbf{100.0} & \textbf{0.4}
& \textbf{100.0} & \textbf{0.0} \\
\midrule

\multirow{5}{*}{GPT5.4-mini}
& Base
& 66.5 & 38.6
& 95.4 & \textbf{0.0} & 95.8 & \textbf{0.0} \\
& SafeJudge
& 61.6 & \textbf{66.4}
& 98.0 & \textbf{0.0} & \textbf{100.0} & \textbf{0.0} \\
& SafePrompt
& 68.2 & 43.1
& 95.2 & \textbf{0.0} & 99.2 & \textbf{0.0} \\
& SafeHarbor
& 66.5 & 45.5
& 98.6 & \textbf{0.0} & 99.4 & \textbf{0.0} \\
\rowcolor{gray!12}
& \textbf{SafeKeep}
& \textbf{72.2} & 57.4
& \textbf{100.0} & \textbf{0.0}
& \textbf{100.0} & \textbf{0.0} \\
\bottomrule
\end{tabular}
\caption{
Evaluation results of SafeKeep and baseline methods across different benchmarks and LLM backends. Metrics with $\uparrow$ ($\downarrow$) indicate that higher (lower) values are better. Refusal, ASR-B, and ASR-E evaluate safety, whereas Acc, Valid-B, and Valid-E evaluate overall task-handling capability. Best results are shown in bold. All values are reported as percentages (\%).
}
\label{tab:main_results}
\end{table*}

\subsection{Benchmarks and Metrics}
We evaluate SafeKeep on two widely adopted benchmarks. 

\begin{itemize}
\item 
\textbf{AgentHarm}~\cite{andriushchenko2025agentharm} contains diverse agent tasks spanning 11 harm categories.
We use its evaluation set, consisting of 176 harmful requests and 176 matched benign requests involving similar tasks and tool specifications. We report accuracy (\textbf{Acc}), defined as the proportion of requests handled correctly, and refusal rate (\textbf{Refusal}), defined as the proportion of harmful requests refused by the agent.

\item \textbf{InjecAgent}~\cite{zhan2024injecagent} evaluates robustness to indirect, observation-level prompt injection using 1,054 attack cases across scenarios. Each case contains a benign user request and a retrieved observation with a malicious instruction, either directly embedded (\emph{basic}) or preceded by an explicit command to ignore prior instructions (\emph{enhanced}). We report the valid rate (Valid), the proportion of outputs that can be parsed as valid actions or responses, and the attack success rate (ASR), the proportion of successful attacks among valid outputs. The corresponding metrics are denoted by \textbf{Valid-B}/\textbf{ASR-B} and \textbf{Valid-E}/\textbf{ASR-E} for the basic and enhanced settings, respectively.

\end{itemize}

Overall, Refusal, ASR-B, and ASR-E measure safety, with higher Refusal and lower ASR indicating better safety; Acc, Valid-B, and Valid-E measure overall task-handling capability, with higher values indicating better capability.

\subsection{Baselines}
We compare SafeKeep against four representative baselines.

\begin{itemize}
\item \textbf{Base} denotes the original agent without any additional safeguard. It serves as the reference point for evaluating each agent's native safety and tool-use capability.

\item \textbf{SafeJudge} uses SafeKeep's two-stage pipeline but retains schema-formatted tool specifications during Safety Judgment, isolating the effect of the flattened representation.

\item \textbf{SafePrompt} appends the AgentHarm safety prompt~\cite{andriushchenko2025agentharm}, which describes common categories of harmful requests and instructs the agent to refuse them, to the input.

\item \textbf{SafeHarbor}~\cite{liusafeharbor} is a recent agent-specific safeguard that adversarially generates context-dependent safety rules, stores them in hierarchical memory, and retrieves relevant rules during inference.
\end{itemize}

\subsection{LLMs}
\label{sec:experimental_setup}.
SafeKeep is model-agnostic and requires no access to model parameters or internal states. We evaluate it using four representative and competitive LLM backends spanning open-source white-box and proprietary black-box models: \textbf{Llama3.1-8B-Instruct}~\cite{grattafiori2024llama}, \textbf{Qwen3-8B}~\cite{yang2025qwen3}, \textbf{Gemini3.1-Flash}~\cite{team2023gemini}, and \textbf{GPT5.4-mini}~\cite{singh2025openai}. These models cover diverse families and providers, allowing us to assess SafeKeep's generalizability across LLM backends.

\subsection{Results}
Table~\ref{tab:main_results} summarizes the evaluation results of SafeKeep.

\paragraph{SafeKeep consistently improves safety across threat settings.}
SafeKeep substantially strengthens safety against both direct harmful requests (i.e., AgentHarm) and indirect, observation-level prompt injection (i.e., InjecAgent). Across the four LLMs, SafeKeep consistently outperforms the unprotected Base agents, increasing the average refusal rate on AgentHarm from 23.8\% to 70.6\%. On InjecAgent, it reduces the average ASR-B and ASR-E from 26.5\% and 24.7\% to 2.3\% and 2.7\%, respectively. Averaging across both injection settings and all four LLMs, the overall attack success rate decreases from 25.6\% to 2.5\%. Overall, SafeKeep achieves the strongest safety performance among the evaluated methods, obtaining the best or tied-best result in 11 of the 12 LLM--safety-metric combinations.

\paragraph{SafeKeep improves safety while largely preserving task-handling capability.}
SafeKeep achieves the highest AgentHarm accuracy across all four LLMs, increasing the average accuracy from 60.9\% for the unprotected Base agents to 79.6\%. Because AgentHarm accuracy jointly rewards correct refusal of harmful requests and correct handling of benign requests, these gains show that SafeKeep does not improve safety merely through indiscriminate refusal. The InjecAgent results exhibit a similar pattern: averaged across the four LLMs, SafeKeep increases Valid-B from 78.7\% to 93.0\% and Valid-E from 78.8\% to 94.8\% relative to the Base agents. 

\paragraph{Safety judgment with flattened textual tool specifications is critical to SafeKeep.}
SafeJudge uses the same two-stage pipeline as SafeKeep but retains schema-formatted tool specifications during Safety Judgment. Across the four LLMs, SafeKeep increases the average refusal rate from 45.0\% to 70.6\% relative to SafeJudge and reduces ASR-B/ASR-E from 19.1\%/22.2\% to 2.3\%/2.7\%.

These results show that self-judgment alone is insufficient; presenting tool specifications in the flattened textual representation is critical for reliable safety assessment.

%% file: sections/9_conclusion.tex
\section{Conclusion}
\label{sec:conclusion}
This paper identifies schema-formatted tool specifications as a source of the safety degradation observed when LLMs are deployed as agents. Through white-box representation analysis, we show that schema formatting induces a hidden-state direction that opposes refusal, weakens refusal-related separation during generation, and causally contributes to harmful tool execution. Building on this finding, we propose SafeKeep, an inference-time safeguard that decouples safety judgment from tool execution by using flattened textual tool specifications for safety assessment while preserving the original agent pipeline for execution. Across two benchmarks and four LLMs, SafeKeep substantially improves agent safety while preserving task-handling capability.